\documentclass[11pt,a4paper]{article}
\usepackage[hyperref]{acl2021}
\usepackage{times}
\usepackage{latexsym}

\usepackage{microtype}

\aclfinalcopy 

\title{Explaining NLP Models via Minimal Contrastive Editing (\mice)}

\author{Alexis Ross\textsuperscript{$\dagger$} \quad Ana Marasovi\'{c}\textsuperscript{$\dagger\diamondsuit$} \quad Matthew E.\ Peters\textsuperscript{$\dagger$} \\ \\
\textsuperscript{$\dagger$}Allen Institute for Artificial Intelligence, Seattle, WA, USA\\
\textsuperscript{$\diamondsuit$}Paul G.\ Allen School of Computer Science and Engineering, University of Washington\\
{\tt\{alexisr,anam,matthewp\}@allenai.org}
}

\date{}
\usepackage{graphicx}
\usepackage{tabularx}
\usepackage{booktabs}
\usepackage{url}
\usepackage{amsmath}
\usepackage{multirow}
\usepackage{float}
\usepackage{booktabs}
\usepackage{amsmath,amssymb,amsthm,xspace}
\usepackage{color,soul}
\usepackage[T1]{fontenc}
\usepackage{comment}
\usepackage{xcolor}
\usepackage{adjustbox}
\usepackage{calc}
\usepackage{subcaption}

\usepackage{hhline}

\usepackage{xspace}

\newcommand{\markedit}[1]{\textbf{\textcolor{red}{#1}}}

\newcommand{\sout}[1]{\st{#1}}

\newcommand{\mattaddressed}[1]{\ignorespaces}

\newcommand{\anaaddressed}[1]{\ignorespaces}
\newcommand\editor{\textsc{Editor}\xspace}
\newcommand\predictor{\textsc{Predictor}\xspace}
\newcommand\predictors{\textsc{Predictors}\xspace}
\newcommand\editors{\textsc{Editors}\xspace}

\newcommand\imdb{\textsc{Imdb}\xspace}
\newcommand\newsgroups{\textsc{Newsgroups}\xspace}
\newcommand\race{\textsc{Race}\xspace}
\newcommand\mice{\textsc{MiCE}\xspace}

\newcommand\randombaseline{\textsc{Rand}\xspace}
\newcommand\micecomplete{\textsc{Grad}\xspace}

\newcommand\highlight{%
  \bgroup
  \expandafter\def\csname sout\space\endcsname{\bgroup \ULdepth =-.8ex \ULset}%
  \markoverwith{\textcolor{yellow}{\rule[-.5ex]{.1pt}{2.5ex}}}%
  \ULon}

\newcommand\sect[1]{\S\ref{#1}}
\usepackage{paralist}

\definecolor{lightorange}{HTML}{FED8B1}

\newlength{\DepthReference}
\newlength{\HeightReference}
\newlength{\Width}

\usepackage{bm}

\usepackage{amsmath}
\DeclareMathOperator{\occurrences}{\text{\#}\_occurrences}
\DeclareMathOperator{\nremovals}{\text{\#}\_removals}
\DeclareMathOperator{\ninsertions}{\text{\#}\_insertions}
\DeclareMathOperator{\totalremovals}{\text{\#}\_all\_removals}
\DeclareMathOperator{\totalinsertions}{\text{\#}\_all\_insertions}
\DeclareMathOperator{\ntokens}{\text{\#}\_all\_tokens}

\begin{document}
\maketitle

\begin{abstract}
Humans have been shown to give contrastive explanations, which explain why an observed event happened \textit{rather than} some other counterfactual event (the \textit{contrast case}). Despite the influential role that contrastivity plays in how humans explain, this property is largely missing from current methods for explaining NLP models. We present \textsc{Minimal Contrastive Editing} (\mice), a method for producing contrastive explanations of model predictions in the form of edits to inputs that change model outputs to the contrast case. Our experiments across three tasks---binary sentiment classification, topic classification, and  multiple-choice question answering---show that \mice is able to produce edits that are not only contrastive, but also \textit{minimal} and \textit{fluent}, consistent with human contrastive edits. We demonstrate how \mice edits can be used for two use cases in NLP system development---debugging incorrect model outputs and uncovering dataset artifacts---and thereby illustrate that producing contrastive explanations is a promising research direction for model interpretability.
\end{abstract}

\section{Introduction}

\begin{figure}[t]
    \centering
    \includegraphics[height=0.26\textheight]{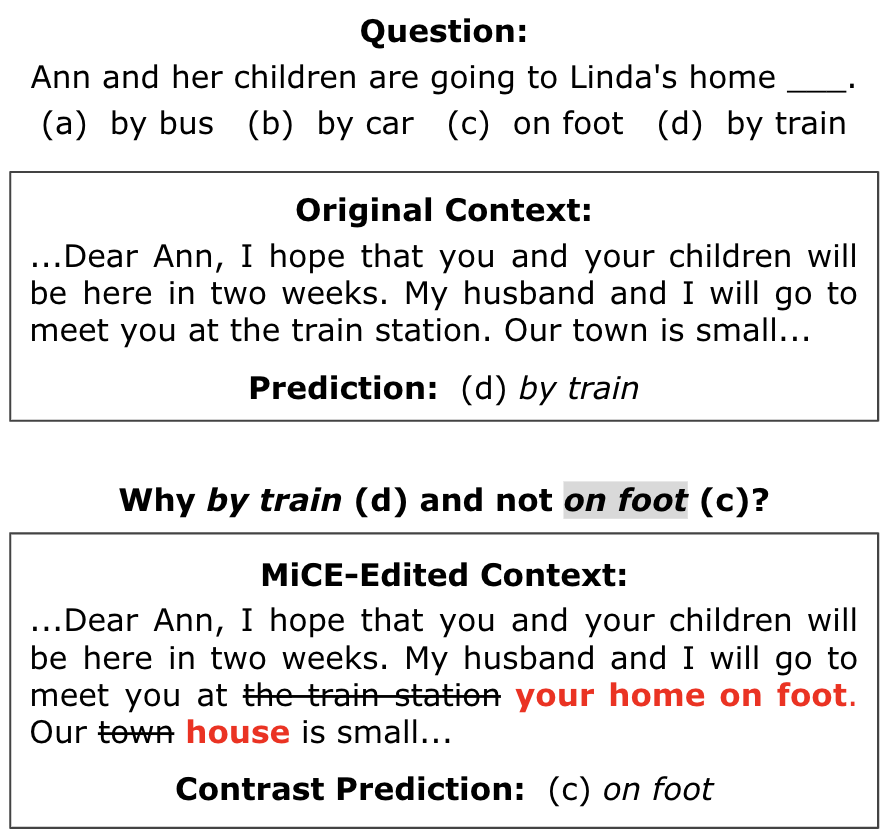}
\caption{An example \mice edit for a multiple-choice question from the \race dataset. \mice generates contrastive explanations in the form of edits to inputs that change model predictions to target (contrast) predictions.
The edit (bolded in red) is minimal and fluent, and it changes the model's prediction from ``by train'' to the contrast prediction ``on foot'' (highlighted in gray).
}
\label{fig:first-example}
\end{figure}

Cognitive science and philosophy research has shown that human explanations are \textbf{contrastive} \citep{Miller2019ExplanationIA}: People explain why an observed event happened rather than some counterfactual event 
called the
\textit{contrast case}. This contrast case plays a key role in modulating what explanations are given. Consider Figure \ref{fig:first-example}.
When we seek an explanation of the model's prediction ``by train,'' we seek it not in absolute terms, but in contrast to another possible prediction (i.e.\ ``on foot'').
Additionally, we tailor our explanation to this contrast case. 
For instance, we might explain why the prediction is ``by train'' and not ``on foot'' by saying that the writer discusses meeting Ann at the train station instead of at Ann's home on foot; such information is captured by the edit (bolded red) that results in the new model prediction ``on foot.'' 
For a different contrast prediction, such as ``by car,'' we would provide a different explanation. In this work, we propose to give contrastive explanations of model predictions in the form of targeted minimal edits, as shown in Figure  \ref{fig:first-example}, that cause the model to change its original prediction to the contrast prediction.

Given the key role that contrastivity plays in human explanations, making model explanations contrastive could make them more user-centered and thus more useful for their intended purposes, such as debugging and exposing dataset biases \cite{Ribera2019CanWD}---purposes which require that \textit{humans} work with explanations \cite{AlvarezMelis2019WeightOE}. However, many currently popular instance-based explanation methods produce highlights---segments of input that support a prediction  \citep{Zaidan2007UsingR,Lei2016RationalizingNP,Chang2019AGT,Bastings2019InterpretableNP,Yu2019RethinkingCR,DeYoung2020ERASERAB,Jain2020LearningTF,10.1162/tacl_a_00254} that can be derived through gradients \citep{Simonyan2014DeepIC,Smilkov2017SmoothGradRN, Sundararajan2017AxiomaticAF}, approximations with simpler models \citep{Ribeiro2016WhySI}, or attention \citep{Wiegreffe2019AttentionIN,Sun2021EffectiveAS}. These methods are not contrastive, as they leave the contrast case undetermined; they do not tell us what would have to be different for a model to have predicted a particular contrast label.\footnote{Free-text rationales \cite{narang2020wt5} can be contrastive if human justifications are collected by asking ``why... instead of...'' which is not the case with current benchmarks \cite{camburu_e_snli_2019, rajani-etal-2019-explain, Zellers2018FromRT}.}

As an alternative approach to NLP model explanation, we introduce \textbf{\textsc{Minimal Contrastive Editing} (\mice)}---a two-stage approach to generating contrastive explanations in the form of targeted minimal edits (as shown in Figure \ref{fig:first-example}).
Given an input, a fixed \predictor model, and a contrast prediction, \mice generates edits to the input that change the \predictor's output from the original prediction to the contrast prediction.  We formally define our edits and describe our approach in \sect{sec:approach}.

We design \mice to produce edits with properties motivated by human contrastive explanations. First, we desire edits to be \textbf{{minimal}}, altering only small portions of input, a property which has been argued to make explanations more intelligible \citep{AlvarezMelis2019WeightOE, Miller2019ExplanationIA}. Second,
\mice edits should be \textbf{{fluent}}, resulting in text natural for the domain and ensuring that any changes in model predictions are not driven by inputs falling out of distribution of naturally occurring text. Our experiments (\sect{sec:evaluation}) on three English-language datasets, \imdb, \newsgroups, and \race, validate that \mice edits are indeed contrastive, minimal, and fluent. 

We also analyze the quality of \mice edits (\sect{sec:analysis}) and show how they may be used for two use cases in NLP system development. First, we show that \mice edits are comparable in size and fluency to human edits on the \imdb dataset. Next, we illustrate how \mice edits can facilitate debugging individual model predictions. Finally, we show how \mice edits can be used to uncover dataset artifacts learned by a powerful \predictor model.\footnote{Our code and trained \editor models are publicly available at \url{https://github.com/allenai/mice}.}

\begin{figure*}[htp]
    \centering
    \includegraphics[height=0.23\textheight]{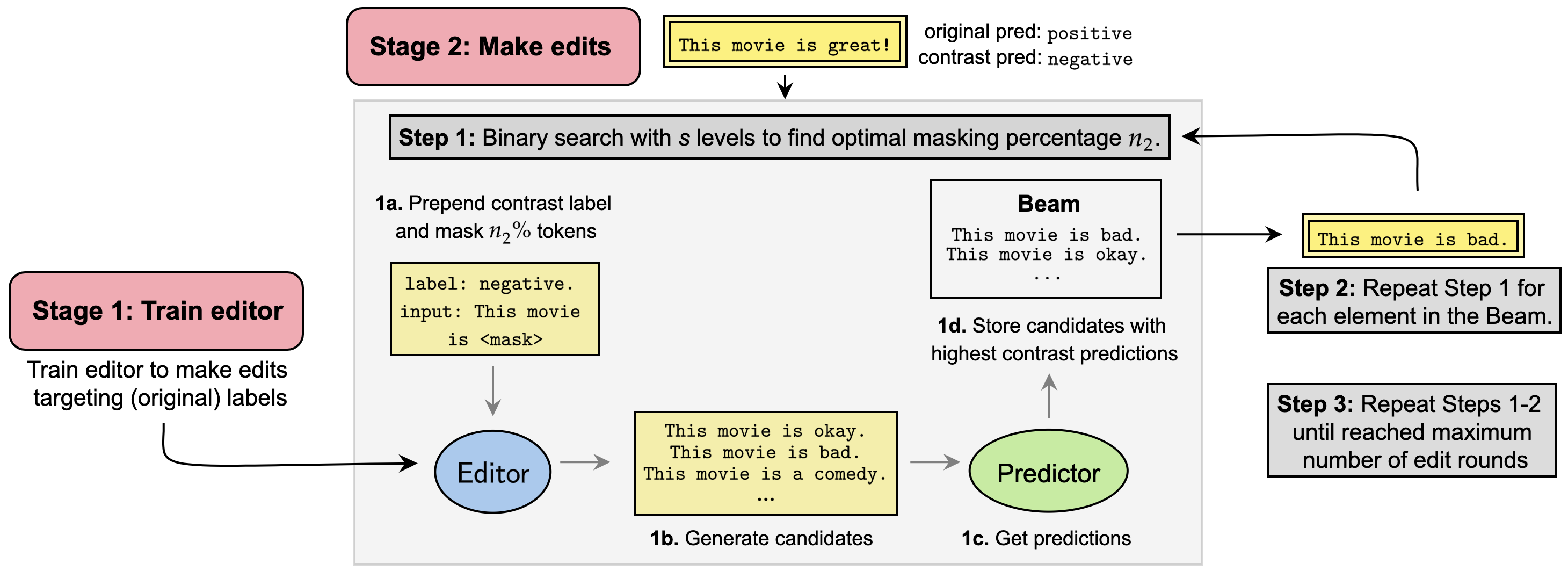}
\caption{An overview of \mice, our two-stage approach to generating edits. In Stage 1 (\sect{ssec:finetuning}), we train the \editor to make edits targeting specific predictions from the \predictor. In Stage 2 (\sect{ssec:editing}), we make contrastive edits with the \editor model from Stage 1 such that the \predictor changes its output to the contrast prediction.}
\label{fig:approach}
\end{figure*}

\section{\mice: Minimal Contrastive Editing}
\label{sec:approach}

This section describes our proposed method, \textsc{Minimal Contrastive Editing}, or \mice, for explaining NLP models with contrastive edits. 

\subsection{\mice Edits as Contrastive Explanations}

Contrastive explanations are answers to questions of the form \textit{Why p and not q?} They explain why the observed event $p$ happened instead of another event $q$, called the \textit{contrast case}.\footnote{Related work also calls it the \textit{foil} \cite{Miller2019ExplanationIA}.} A long line of research in the cognitive sciences and philosophy has found that human explanations are contrastive \citep{VanFraassen1980,Lipton1990, Miller2019ExplanationIA}. Human contrastive explanations have several hallmark characteristics. 
First, they cite \textit{contrastive features}: features that result in the contrast case when they are changed in a particular way \citep{ChinParker2017ContrastiveCG}. %
Second, they are minimal in the sense that they rarely cite the entire causal chain of a particular event, but select just a few relevant causes \citep{Hilton2017}. In this work, we argue that a minimal edit to a model input that causes the model output to change to the contrast case has both these properties and can function as an effective contrastive explanation. We first give an illustration of contrastive explanations humans might give and then show how minimal contrastive edits offer analogous contrastive information.

As an example, suppose we want to explain why the answer to the question ``\textit{Q}: Where can you find a clean pillow case that is not in use?'' 
is ``\textit{A}: the drawer.''\footnote{Inspired by an example in \citet{Talmor2019CommonsenseQAAQ}: Question: ``Where would you store a pillow case that is not in use?'' Choices: ``drawer, kitchen cupboard, bedding store, england.''}
If someone asks why the answer is not ``\textit{C1}: on the bed,'' we might explain: ``\textit{E1}: Because only the drawer stores pillow cases that are not in use.''
However, \textit{E1} would \textit{not} be an explanation of why the answer is not ``\textit{C2}: in the laundry hamper,'' since both drawers and laundry hampers store pillow cases that are not in use. For contrast case \textit{C2}, we might instead explain: ``\textit{E2}: Because only laundry hampers store pillow cases that are not clean.'' We cite different parts of the original question depending on the contrast case.

In this work, we propose to offer contrastive explanations in the form of minimal edits that result in the contrast case as model output. Such edits are effective contrastive explanations because, by construction, they highlight contrastive features.
For example, a contrastive edit of the original question for contrast case \textit{C1} would be: 
``Where can you find a clean pillow case that is \textbf{\sout{not}} in use?''; the information provided by this edit---that it is whether or not the pillow case is in use that determines whether the answer is ``the drawer'' or ``on the bed''---is analogous to the information provided by \textit{E1}. Similarly, a contrastive edit for contrast case \textit{C2} that changed the question to ``Where can you find a \textbf{\sout{clean} dirty} pillow case that is not in use?'' provides analogous information to \textit{E2}.

\subsection{Overview of \mice}
\label{ssec:overview}
We define a contrastive edit to be a modification of an input instance 
that causes a \predictor model (whose behavior is being explained) to change its output from its original prediction for the unedited input 
to a given target (contrast) prediction. Formally, for textual inputs, given a fixed \predictor $f$, input $\boldsymbol{x}=(x_1, x_2, ..., x_N)$ of $N$ tokens, original prediction $f(\boldsymbol{x})=y_p$ and contrast prediction $y_c \neq y_p$, a contrastive edit is a mapping $e: (x_1, ..., x_N) \rightarrow (x_1', ..., x_M')$ such that $f(e(\boldsymbol{x}))=y_c$.

We propose \mice, a two-stage approach to generating contrastive edits, illustrated in Figure \ref{fig:approach}. In Stage 1, we prepare a highly-contextualized \editor model to associate edits with given 
end-task labels (i.e., labels for the task of the \predictor) such that the contrast label $y_c$ is not ignored in \mice's second stage. Intuitively, we do this by masking the spans of text that are ``important'' for the given target label (as measured by the \predictor's gradients) and training our \editor to reconstruct these spans of text given the masked text and target label as input. In Stage 2 of \mice, we generate contrastive edits $e(\boldsymbol{x})$ using the \editor model from Stage 1. Specifically, we generate candidate edits $e(\boldsymbol{x})$ by masking different percentages of $\boldsymbol{x}$ and giving masked inputs with prepended contrast label $y_c$ to the \editor; we use binary search to find optimal masking percentages and beam search to keep track of candidate edits that result in the highest probability of the contrast labels $p(y_c|e(\boldsymbol{x}))$ given by the \predictor.

\subsection{Stage 1: Fine-tuning the \editor}
\label{ssec:finetuning}

In Stage 1 of \mice, we fine-tune the \editor to infill masked \textit{spans} of text in a targeted manner. Specifically, we fine-tune a pretrained model to infill masked spans given masked text and a target end-task label as input. In this work, we use the \textsc{Text-to-Text Transfer Transformer (T5)} model \citep{Raffel2019ExploringTL} as our pretrained \editor, but any model suitable for span infilling can in principle be the \editor in \mice. The addition of the target label allows the highly-contextualized \editor to condition its predictions on both the masked context and the given target label such that the contrast label is not ignored in Stage 2. What to use as target labels during Stage 1 depends on who the end-users of \mice are.
The end-user could be: (1) a model developer who has access to the labeled data used to train the predictor, or (2) lay-users, domain experts, or other developers 
without
access to the labeled data. In the former case, we could use the gold label as targets, and in the latter case, we could use the labels predicted by \predictor. Therefore, during fine-tuning, we experiment with using both gold labels and original predictions $y_p$ of our \predictor model as target labels. To provide target labels, we prepend them to inputs to the \editor. For more information about how these inputs are formatted, see Appendix \ref{sec:data-processing}. Results in Table \ref{tab:no-label} show that fine-tuning with target labels results in better edits than fine-tuning without them.

The above procedure allows our \editor to condition its infilled spans on both the context and the target label. But this still leaves open the question of where to mask our text. Intuitively, we want to mask the tokens that contribute most to the \predictor's predictions, since these are the tokens that are most strongly associated with the target label. We propose to use gradient attribution \citep{Simonyan2014DeepIC} to choose tokens to mask. For each instance, we take the gradient of the predicted logit for the target label with respect to the embedding layers of $f$ and take the $\ell_1$ norm across the embedding dimension. We then mask the $n_1\%$ of tokens with the highest gradient norms. We replace consecutive tokens (i.e.,\ spans) with sentinel tokens, following \citet{Raffel2019ExploringTL}. Results in Table \ref{tab:main-results} show that gradient-based masking outperforms random masking.

\subsection{Stage 2: Making Edits with the \editor}
\label{ssec:editing}

In the second stage of our approach, we use our fine-tuned \editor to make edits using beam search \cite{reddy1977speech}. In each round of edits, we mask consecutive spans of $n_2\%$ of tokens in the original input, prepend the contrast prediction to the masked input, and feed the resulting masked instance to the \editor; the \editor then generates $m$ edits. The masking procedure during this stage is gradient-based as in Stage 1. 

In one round of edits, we conduct a binary search with $s$ levels over values of $n_2$ between values $n_2 = 0\%$ to $n_2 = 55\%$ to efficiently find a value of $n_2$ that is large enough to result in the contrast prediction while also modifying only minimal parts of the input. After each round of edits, we get $f$'s predictions on the edited inputs, order them by contrast prediction probabilities, and update the beam to store the top $b$ edited instances. As soon as an edit $e^{*}=e(\boldsymbol{t})$ is found that results in the contrast prediction, i.e.,\ $f(e^{*}) = y_c$, we stop the search procedure and return this edit. For generation, we use a combination of top-\textit{k} \citep{Fan2018HierarchicalNS} and top-\textit{p} (nucleus) sampling \citep{Holtzman2020TheCC}.\footnote{We use this combination because we observed in preliminary experiments that it led to good results.}

\begin{table*}[htbp]
\centering
\small
\begin{tabular}{lccc|ccc|ccc}
\toprule
\multirow{3}{*}{\large\begin{tabular}{@{}c@{}}\large\textbf{\mice} \\ \textsc{Variant}\end{tabular}}& \multicolumn{3}{c|}{\textbf{\large\imdb}} & \multicolumn{3}{c|}{\textbf{\large\newsgroups}} & \multicolumn{3}{c}{\textbf{\large\race}}\\

& \footnotesize $\uparrow$ &  \footnotesize $\downarrow$ &  \footnotesize $\approx1$ & \footnotesize $\uparrow$ &  \footnotesize $\downarrow$ &  \footnotesize $\approx1$& \footnotesize $\uparrow$ &  \footnotesize $\downarrow$ &  \footnotesize $\approx1$\\

& \small \textbf{Flip Rate} &  \small \textbf{Minim.} &  \small \textbf{Fluen.} &  \small \textbf{Flip Rate} &  \small \textbf{Minim.} &  \small \textbf{Fluen.} &  \small \textbf{Flip Rate} &  \small \textbf{Minim.} &  \small \textbf{Fluen.}\\

\midrule\midrule
\normalsize\textbf{\textsc{{*Pred + Grad}}} \small & $\mathbf{1.000}$ & $\mathbf{0.173}$ & $\mathbf{0.981}$ & $\mathbf{0.992}$ & $\mathbf{0.261}$ & $0.968$ & $0.915$ & $\mathbf{0.331}$ & $\mathbf{0.981}$\\

\normalsize\textbf{\textsc{*Gold + Grad}} & $\mathbf{1.000}$  & $0.185$ & $0.979$ & $\mathbf{0.992}$ & ${0.271}$ & $0.966$ & $\mathbf{0.945}$ & $0.335$ & $0.979$\\
\midrule
\normalsize\textsc{Pred + }\normalsize\randombaseline \small & $\mathbf{1.000}$ & $0.257$ & $0.958$ & $0.968$ & $0.378$ & $0.928$ & $0.799$ & $0.440$ & $0.953$\\

\normalsize\textsc{Gold + }\randombaseline & $\mathbf{1.000}$ & $0.302$ & $0.952$ & $0.965$ & $0.370$ & $0.929$ & $0.801$ & $0.440$ & $0.955$\\

\midrule
\normalsize \textsc{No-Finetune} \small & $0.995$ & $0.360$ & $0.960$ & $0.941$ & $0.418$ & $0.938$ & -- & -- & --\\

\bottomrule
\end{tabular}
\caption{Efficacy of the \mice procedure. We evaluate \mice edits on three metrics (described in \sect{ssec:experimental-setup}): {flip rate}, {minimality}, and {fluency}. We report mean values for minimality and fluency.
* marks full \mice variants; others explore ablations. For each property (i.e., column), the best value across \mice variants is bolded. We experiment with \predictor's predictions (\textsc{Pred}) and gold labels (\textsc{Gold}) as target labels during Stage 1. Across datasets, our \textsc{Grad} \mice procedure achieves a high flip rate with small and fluent edits.}
\label{tab:main-results}
\end{table*}

\section{Evaluation}

\label{sec:evaluation}

This section presents empirical findings that \mice produces minimal and fluent contrastive edits.

\subsection{Experimental Setup}
\label{ssec:experimental-setup}

\paragraph{Tasks} We evaluate \mice on three English-language datasets: {\textsc{IMDB}}, a binary sentiment classification task \citep{Maas2011LearningWV}, a 6-class version of the {\textsc{20 Newsgroups}} topic classification task \citep{Lang95}, and {\race}, a multiple choice question-answering task \citep{Lai2017RACELR}.\footnote{We create this 6-class version by mapping the 20 existing subcategories to their respective larger categories---i.e. ``talk.politics.guns'' and ``talk.religion.misc'' $\rightarrow$ ``talk.'' We do this in order to make the label space smaller. The resulting classes are: alt, comp, misc, rec, sci, and talk.}

\paragraph{\textbf{\predictors}} \mice can be used to make contrastive edits for any differentiable \predictor model, i.e., any end-to-end neural model.  In this paper, for each task, we train a \predictor model $f$ built on \textsc{RoBERTa-large} \citep{Liu2019RoBERTaAR}, and fix it during evaluation. The test accuracies of our \predictors are 95.9\%, 85.3\% and 84\% for \imdb, \newsgroups, and \race, respectively. For training details, see Appendix \ref{ssec:training-details}.

\paragraph{\textbf{\editors}} Our \editors build on the base version of \textsc{T5}. For fine-tuning our \editors (Stage 1), we use the original training data used to train \predictors. We randomly split the data, 75\%/25\% for fine-tuning/validation and fine-tune until the validation loss stops decreasing (for a max of 10 epochs) with $n_1\%$ of tokens masked, where $n_1$ is a randomly chosen value in $[20,55]$. For more details, see Appendix \ref{ssec:finetuning-details}. In Stage 2, for each instance, we set the label with the second highest predicted probability as the contrast prediction. We set beam width $b = 3$, consider $s = 4$ search levels during binary search over $n_2$ in each edit round, 
and run our search for a max of 3 edit rounds. 
For each $n_2$, we sample $m = 15$ generations from our fine-tuned \editors with $p=0.95$, $k=30$.
\footnote{We tune these hyperparameters on a $50$-instance subset of the \imdb validation set prior to evaluation. We note that for larger values of $n_2$, the generations produced by the \textsc{T5} \editors sometimes degenerate; see Appendix \ref{sec:appendix_t5_long} for details.}

\paragraph{Metrics}
We evaluate \mice on the test sets of the three datasets. The \race and \newsgroups test sets contain 4,934 and 7,307 instances, respectively.\footnote{For the \newsgroups test set, there are 7,307 instances remaining after filtering out empty strings.}  For \imdb, we randomly sample 5K of the 25K instances in the test set for evaluation because of the computational demands of evaluation.
\footnote{A single contrastive edit is expensive and takes an average of $\approx 15$ seconds per \imdb instance ($\approx 230$ tokens). Calculating the fluency metric adds an additional average of $\approx 16.5$ seconds per \imdb instance. For more details, see Section \ref{sec:discussion}.}

For each dataset, we measure the following three properties: (1)\,\,\textbf{flip rate:} the proportion of instances for which an edit results in the contrast label; (2)\,\,\textbf{minimality:} the ``size'' of the edit as measured by the word-level Levenshtein distance between the original and edited input, which is the minimum number of deletions, insertions, or substitutions required to transform one into the other. We report a normalized version of this metric with a range from 0 to 1---the Levenshtein distance divided by the number of words in the original input; (3)\,\, \textbf{fluency:} a measure of how similarly distributed the edited output is to the original data.
We evaluate fluency by comparing masked language modeling loss on both the original and edited inputs using a pretrained model. Specifically, given the original $N$-length sequence, we create $N$ copies, each with a different token replaced by a mask token, following \citet{Salazar2020MaskedLM}. We then take a pretrained \textsc{t5-base} model and compute the average loss across these $N$ copies. We compute this loss value for both the original input and edited input and report their \textit{ratio}---i.e., edited $/$ original. We aim for a value of 1.0, which indicates equivalent losses for the original and edited texts. When \mice finds multiple edits, we report metrics for the edit with the smallest value for minimality.

\subsection{Results}
\label{ssec:results}

Results are shown in Table \ref{tab:main-results}. Our proposed  \micecomplete \mice procedure (upper part of Table \ref{tab:main-results}) achieves a high flip rate across all three tasks. This is the outcome regardless of whether predicted target labels (first row, 91.5--100\% flip rate) or gold target labels (second row, 94.5--100\% flip rate) are used for fine-tuning in Stage 1. We observe a slight improvement from using the gold labels for the \race \predictor, which may be explained by the fact that it is less accurate (with a training accuracy of $89.9\%$) than the \imdb and \newsgroups classifiers.

\mice achieves a high flip-rate while its edits remain small and result in fluent text.
In particular, \mice on average changes 17.3--33.1\% of the original tokens when predicted labels are used in Stage 1 and 18.5--33.5\% with gold labels. Fluency is close to 1.0 indicating no notable change in mask language modeling loss after the edit---i.e., edits fall in distribution of the original data. We achieve the best results across metrics on the \imdb dataset, as expected since \imdb is a binary classification task with a small label space. These results demonstrate that \mice presents a promising research direction for the generation of contrastive explanations; however, there is still room for improvement, especially for more challenging tasks such as \race.

In the rest of this section, we provide results from several ablation experiments.

\paragraph{\textbf{Fine-tuning vs.\ No Fine-tuning}} We investigate the effect of fine-tuning (Stage 1) with a baseline that skips Stage 1 altogether. For this \textsc{No-Finetune} baseline variant of \mice, we use the vanilla pretrained \textsc{T5-base} as our \editor. As shown in Table \ref{tab:main-results}, the \textsc{No-Finetune} variant underperforms all other (two-stage) variants of \mice for the \imdb and \newsgroups datasets.\footnote{We leave \race out from our evaluation with the \textsc{No-Finetune} baseline because we observe that the pretrained \textsc{T5} model does not generate text formatted as span infills; we hypothesize that this model has not been trained to generate infills for masked inputs formatted as multiple choice inputs.} Fine-tuning particularly improves the minimality of edits, while leaving the flip rate high. We hypothesize that this effect is due to the effectiveness of Stage 2 of \mice at finding contrastive edits: Because we iteratively generate many candidate edits using beam search, we are likely to find a prediction-flipping edit. Fine-tuning allows us to find such an edit at a lower masking percentage.

\paragraph{\textbf{Gradient vs.\ Random Masking}} We study the impact of using gradient-based masking in Stage 1 of the \mice procedure with a \randombaseline variant, which masks spans of randomly chosen tokens. As shown in the middle part of Table \ref{tab:main-results}, gradient-based masking outperforms random masking when using both predicted and gold labels across all three tasks and metrics, suggesting that the gradient-based attribution used to mask text during Stage 1 of \mice is an important part of the procedure. The differences are especially notable for \race, which is the most challenging task according to our metrics.

\begin{table}
\centering
\small
\begin{tabular}{ll ccc}
\toprule
\multicolumn{2}{l}{\normalsize\textbf{\imdb Condition}} & \footnotesize $\uparrow$ &  \footnotesize $\downarrow$ &  \footnotesize $\approx 1$\\

\normalsize\textbf{Stage 1} & \normalsize\textbf{Stage 2} & \small \textbf{Flip Rate} &  \small \textbf{Minim.} &  \small \textbf{Fluen.}\\

\midrule\midrule
No Label & No Label & $0.994$ & $0.369$ & $0.966$\\
No Label & Label & $0.997$ & $0.362$ & $0.967$\\
Label & No Label & $0.999$ & $0.327$ & $0.968$\\
\midrule
Label &  Label & $\mathbf{1.000}$ & $\mathbf{0.173}$ & $\mathbf{0.981}$\\

\bottomrule
\end{tabular}
\caption{Effect of using target end-task labels during the two stages of \textsc{PRED}+\textsc{GRAD} \mice on the \imdb dataset. When end-task labels are provided, they are original \predictor labels during Stage 1 and contrast labels during Stage 2. The best values for each property (column) are bolded. Using end-task labels during both Stage 1 (\editor fine-tuning) and Stage 2 (making edits) of \mice outperforms all other conditions.}
\label{tab:no-label}
\end{table}

\paragraph{\textbf{Targeted vs. Un-targeted Infilling}} We investigate the effect of using target labels in both stages of \mice by experimenting with removing target labels during Stage 1 (\editor fine-tuning) and Stage 2 (making edits). As shown in Table \ref{tab:no-label}, we observe that giving target labels to our \editors during both stages of 
\mice  improves edit quality. Fine-tuning \editors without labels in Stage 1 (``No Label'') leads to worse flip rate, minimality, and fluency than does fine-tuning \editors with labels (``Label''). Minimality is particularly affected, and we hypothesize that using target end-task labels in both stages provides signal that allows the \editor in Stage 2 to generate prediction-flipping edits at lower masking percentages.

\newcolumntype{M}[1]{>{\centering\arraybackslash}m{#1}}

\begin{table*}[htbp]
\centering
\footnotesize
\begin{tabular}{M{0.05\linewidth}| m{0.85\linewidth}}\toprule
\imdb & \begin{center}\textbf{Original pred} $y_p = $ {\underline{positive}} \, \, \, \textbf{Contrast pred} $y_c = $ {negative}\end{center}
An interesting pairing of stories, this little flick manages to bring together seemingly different characters and story lines all in the backdrop of WWII and succeeds in tying them together without losing the audience. I was impressed by the depth portrayed by the different characters and also by how much I really felt I understood them and their motivations, even though the time spent on the development of each character was very limited. The outstanding acting abilities of the individuals involved with this picture are easily noted. A fun, stylized movie with a slew of comic moments and a bunch more head shaking events. \sout{7/10} \markedit{4/10} \\
\midrule

\multirow{10}{*}{\race} & \multicolumn{1}{c}{\begin{tabular}{m{0.75\linewidth}}\vspace{-0.5mm} \begin{center}\hspace{1 mm} \textbf{Question:} \hspace{0.5 mm} Mark went up in George's plane \, \underline{\hspace{2cm}}. \\
\hspace{1 mm} \hspace{0.5 mm} (a)\, {twice} \, {(b)\, \underline{only once}} \,(c)\, {several times} \, (d)\, {once or twice}.\end{center}
\end{tabular}}\vspace{-3.5mm}\\
& \begin{center}\textbf{Original pred} $y_p = (a)\,$ {twice} \, \, \, \textbf{Contrast pred} $y_c = (b)\,$ {\underline{only once}}\end{center} When George was thirty-five, he bought a small plane and learned to fly it. He soon became very good and made his plane do all kinds of tricks. George had a friend, whose name was Mark. One day George offered to take Mark up in his plane. Mark thought, "I've traveled in a big plane several times, but I've never been in a small one, so I'll go." They went up, and George flew around for half an hour and did all kinds of tricks in the air. When they came down again, Mark was glad to be back safely, and he said to his friend in a shaking voice, "Well, George, thank you very much for those two \sout{trips} \markedit{tricks} in your plane." George was very surprised and said, "Two \sout{trips?} \markedit{tricks.}" \sout{Yes,} \markedit{That's} my first \sout{and my last} \markedit{time, George}." \sout{answered} \markedit{said} Mark.\\

\bottomrule
\end{tabular}
\caption{Examples of edits produced by \mice. Insertions are bolded in red. Deletions are struck through. $y_p$ is the \predictor's original prediction, and $y_c$ the contrast prediction. True labels for original inputs are underlined.}
\label{tab:examples}
\end{table*}

\section{Analysis of Edits}
\label{sec:analysis}

In this section, we compare \mice edits with human contrastive edits. Then, we turn to a key motivation for this work: the potential for contrastive explanations to assist in NLP system development. We show how \mice edits can be used to debug incorrect predictions and uncover dataset artifacts.

\subsection{Comparison with Human Edits}
\label{ssec:human}

We ask whether the contrastive edits produced by \mice are minimal and fluent in a meaningful sense. In particular, we compare these two metrics for \mice edits and human contrastive edits. We work with the \imdb contrast set created by \citet{Gardner2020EvaluatingNM}, which consists of original test inputs and human-edited inputs that cause a change in \textit{true} label. We report metrics on the subset of this contrast set for which the human-edited inputs result in a change in model prediction for our \imdb \predictor; this subset consists of $76$ instances. The flip rate of \mice edits on this subset is 100\%. The mean minimality values of human and \mice edits are $\mathbf{0.149}$ (human) and $\mathbf{0.179}$ (\mice), and the mean fluency values are $\mathbf{1.01}$ (human) and $\mathbf{0.949}$ (\mice). The similarity of these values suggests that \mice edits are comparable to human contrastive edits along these dimensions. 

We also ask to what extent human edits overlap with \mice edits. For each input, we compute the overlap between the original tokens changed by humans and the original tokens edited by \mice. The mean number of overlapping tokens,
normalized by the number of original tokens edited by humans, is $\mathbf{0.298}$. Thus, while there is some overlap between \mice and human contrastive edits, they generally change different parts of text.\footnote{\mice edits explain \predictors' behavior and therefore need not be similar to human edits, which are designed to change gold labels.} This analysis suggests that there may exist multiple informative contrastive edits for a single input. Future work can investigate and compare the different kinds of insight that can be obtained through human and model-driven contrastive edits.

\subsection{Use Case 1: Debugging Incorrect Outputs}

Here, we illustrate how \mice edits can be used to debug incorrect model outputs. Consider the \race input in Table \ref{tab:examples}, for which the \race \predictor gives an incorrect prediction. In this case, a model developer may want to understand why the model got the answer wrong. This setting naturally brings rise to a contrastive question, i.e., \textit{Why did the model predict the wrong choice (``twice'') instead of the correct one (``only once'')?} 

The \mice edit shown offers insight into this question: Firstly, it highlights which part of the paragraph has an influence on the model prediction---the last few sentences. Secondly, it reveals that a source of confusion is Mark's joke about having traveled in George's plane twice, as changing Mark's dialogue from talking about a ``first and...last'' trip to a single trip results in a correct model prediction.

\mice edits can also be used to debug model \textit{capabilities} by offering hypotheses about ``bugs'' present in models: For instance, the edit in Table \ref{tab:examples} might prompt a developer to investigate whether this \predictor lacks non-literal language understanding capabilities. In the next section, we show how insight from individual \mice edits can be used to uncover a bug in the form of a dataset-level artifact learned by a model. In Appendix \ref{sec:buggy_case_study}, we further analyze the debugging utility of \mice edits with a \predictor \textit{designed} to contain a bug.

\begin{table}[]
\centering
\small
\begin{tabular}{c@{\hspace{1.25\tabcolsep}}c|c@{\hspace{1.25\tabcolsep}}c}
\toprule
\multicolumn{2}{c|}{\normalsize $y_c=$ \textit{\textbf{positive}}} & \multicolumn{2}{c}{\normalsize $y_c=$ \textit{\textbf{negative}}}\\
\textbf{Removed} & \textbf{Inserted} & \textbf{Removed} & \textbf{Inserted}\\\midrule
4/10 & excellent & 10/10 & awful\\
ridiculous & enjoy & 8/10 & disappointed\\
horrible & amazing & 7/10 & 1\\
4 & entertaining & 9 & 4\\
predictable & 10 & enjoyable & annoying\\

\bottomrule
\end{tabular}
\caption{Top 5 \imdb tokens edited by \mice at a higher rate than expected given their original frequency (\sect{ssec:artifacts}). Results are separated by contrast predictions.}
\label{tab:artifacts}
\end{table}

\subsection{Use Case 2: Uncovering Dataset Artifacts}
\label{ssec:artifacts}
Manual inspection of some edits for \imdb suggests that the \imdb \predictor has learned to rely heavily on numerical ratings. For instance, in the \imdb example in Table \ref{tab:examples}, the \mice edit results in a negative prediction from the \predictor even though the edited text is overwhelmingly positive. We test this hypothesis by investigating whether numerical tokens are more likely to be edited by \mice. 

We analyze the edits produced by \mice (\textsc{Gold + Grad}) described in  \sect{ssec:experimental-setup}. We limit our analysis to a subset of the 5K instances for which the edit produced by \mice has a minimality value of $\leq$0.05, as we are interested in finding simple artifacts driving the predictions of the \imdb \predictor; this subset has 902 instances. We compute three metrics for each unique token, i.e., type $t$: 
{
\setlength{\abovedisplayskip}{6pt}
  \setlength{\belowdisplayskip}{\abovedisplayskip}
  \setlength{\abovedisplayshortskip}{0pt}
  \setlength{\belowdisplayshortskip}{3pt}
\begin{align*}
    p(t) &= \occurrences(t) / \ntokens,\\
    p_r(t) &= \nremovals(t) / \totalremovals,\\
   p_i(t) &= \ninsertions(t) / \totalinsertions,
\end{align*}
}%
and report the tokens with the highest values for the ratios $p_r(t)/p(t)$ and $p_i(t)/p(t)$. Intuitively, these tokens are removed/inserted at a higher rate than expected given the frequency with which they appear in the original \imdb inputs. We exclude tokens that occur $<$10 times from our analysis.

Results from this analysis are shown in Table \ref{tab:artifacts}. In line with our hypothesis, we observe a bias towards removing low numerical ratings and inserting high ratings when the contrast prediction $y_c$ is positive, and vice versa when $y_c$ is negative. In other words, in the presence of a numerical score, the \predictor may ignore the content of the review and base its prediction solely on the score (as in the \imdb example in Table 3).

\section{Discussion}
\label{sec:discussion}

In this section, we reflect on \mice's shortcomings. Foremost, \mice is computationally expensive. Stage 1 requires fine-tuning a large pretrained generation model as the \editor. 
More significantly, Stage 2 requires multiple rounds of forward and backward passes to find a minimal edit: Each edit round in Stage 2 requires $b \times s \times m$ decoded sequences with the \editor, as well as $b \times s \times m$ forward passes and $b$ backward passes with the \predictor (with $b = 1$ the first edit round), where $b$ is the beam width, $s$ is the number of search levels in binary search over the masking percentages, and $m$ is the number of generations sampled for each masking percentage.
Our experiments required 180 forward passes, 180 decoded sequences, and 3 backward passes for edit rounds after the first.

While efficient search for targeted edits is an open challenge in other fields of machine learning \citep{Russell2019EfficientSF, Dandl2020MultiObjectiveCE}, this problem is even more challenging for language data, as the space of possible perturbations is much larger than for tabular data. An important future direction is to develop more efficient methods of finding edits.

This shortcoming prevents us from finding edits that are minimal in a precise sense. In particular, we may be interested in a constrained notion of minimality that defines an edit $e(\boldsymbol{x})$ as minimal if there exists no subset of $e(\boldsymbol{x})$ that results in the contrast prediction. Future work might consider creating methods to produce edits with this property.

\section{Related Work}

The problem of generating minimal contrastive edits, also called counterfactual explanations \cite{Wachter2017CounterfactualEW},\footnote{Formally, methods for producing targeted counterfactual explanations solve the same task as \mice. However, not all contrastive explanations are counterfactual explanations; contrastive explanations can take forms beyond contrastive edits, such as free-text rationales \cite{Liang2020ALICEAL} or highlights \cite{Jacovi2020AligningFI}. In this paper, we choose to refer to \mice edits as ``contrastive'' rather than ``counterfactual'' because we seek to argue for the utility of \textit{contrastive explanations} of model predictions more broadly; we present \mice as one method for producing contrastive explanations of a particular form and hope future work will explore different forms of contrastive explanations.} has previously been explored for tabular data \cite{Karimi2020ModelAgnosticCE} and images \cite{Hendricks2018GroundingVE, Goyal2019CounterfactualVE, Looveren2019InterpretableCE} but less for language. Recent work explores the use of minimal edits changing true labels for evaluation \citep{Gardner2020EvaluatingNM} and data augmentation \citep{Kaushik2020LearningTD, Teney2020LearningWM}, whereas we focus on minimal edits changing model predictions for \textit{explanation}. 

\paragraph{Contrastive Explanations within NLP} 
There exist limited methods for automatically generating contrastive explanations of NLP models.
\citet{Jacovi2020AligningFI} define contrastive highlights, which are determined by the inclusion of contrastive features; in contrast, our contrastive edits specify \textit{how} to edit (vs. whether to include) features and can insert new text.\footnote{See Appendix \ref{sec:buggy_case_study} for a longer discussion about the advantage of inserting new text in explanations, which \mice edits can do but methods that attribute feature importance (i.e.\, highlights) cannot.} \citet{li-etal-2020-linguistically} generate counterfactuals using linguistically-informed transformations (\textsc{LIT}), and \citet{Yang2020GeneratingPC} generate counterfactuals for binary financial text classification using grammatically plausible single-word edits (\textsc{REP-SCD}). Because both methods rely on manually curated, task-specific rules, they cannot be easily extended to tasks without predefined label spaces, such as \race.\footnote{\textsc{LIT} relies on hand-crafted transformation for NLI tasks based on linguistic knowledge, and \textsc{REP-SCD} makes antonym-based edits using manually curated, domain-specific lexicons for each label.} Most recently, \citet{Jacovi2021ContrastiveEF} propose a method for producing contrastive explanations in the form of latent representations; in contrast, \mice edits are made at the textual level and are therefore more interpretable. 

This work also has ties to the literature on causal explanation \cite{CausalityPearl}. Recent work within NLP derives causal explanations of models through counterfactual interventions \cite{CausalLM, GenderBiasCausal}. The focus of our work is the largely unexplored task of creating targeted interventions for language data; however, the question of how to derive causal relationships from such interventions remains an interesting direction for future work.

\paragraph{Counterfactuals Beyond Explanations}
Concurrent work by \citet{Madaan2020GenerateYC} applies controlled text generation methods to generate targeted counterfactuals and explores their use as test cases and augmented examples in the context of classification. Another concurrent work by \citet{Wu2021PolyjuiceAG} presents \textsc{Polyjuice}, a general-purpose, untargeted counterfactual generator. Very recent work by \citet{Sha2021ControllingTE}, introduced after the submission of \mice, proposes a method for targeted contrastive editing for Q\&A that selects answer-related tokens, masks them, and generates new tokens. Our work differs from these works in our novel framework for efficiently finding \textit{minimal} edits (\mice Stage 2) and our use of edits as explanations.

\paragraph{Connection to Adversarial Examples}
Adversarial examples are minimally edited inputs that cause models to incorrectly change their predictions despite no change in true label \cite{Jia2017AdversarialEF, Ebrahimi2018HotFlipWA, Pal2020ToTO}. Recent methods for generating adversarial examples also preserve fluency \cite{Zhang2019GeneratingFA, Li2020BERTATTACKAA, song-etal-2020-adversarial}\footnote{\citet{song-etal-2020-adversarial} propose a method to produce fluent \textit{semantic collisions}, which they call the ``inverse'' of adversarial examples.}; however, adversarial examples are designed to find \textit{erroneous} change in model outputs; contrastive edits place no such constraint on model correctness. Thus, current approaches to generating adversarial examples, which can exploit semantics-preserving operations \citep{Ribeiro2018SemanticallyEA} such as paraphrasing \citep{iyyer-etal-2018-adversarial} or word replacement \citep{alzantot-etal-2018-generating, Ren2019GeneratingNL, Garg2020BAEBA}, cannot be used to generate contrastive edits. 

\paragraph{Connection to Style Transfer} 
The goal of style transfer is to generate minimal edits to inputs to result in a target style (sentiment, formality, etc.) \citep{Fu2018StyleTI,Li2018DeleteRG, Goyal2020MultidimensionalST}. 
Most existing approaches train an encoder to learn style-agnostic latent representation of inputs and train attribute-specific decoders to generate text reflecting the content of inputs but exhibiting a different target attribute \citep{Fu2018StyleTI,Li2018DeleteRG, Goyal2020MultidimensionalST}. 
Recent works by \citet{Wu2019MaskAI} and \citet{Malmi2020UnsupervisedTS} adopt two-stage approaches that first identify where to make edits and then make them using pretrained language models. Such approaches can only be applied to generate contrastive edits for classification tasks with well-defined “styles,” which exclude more complex tasks such as question answering. 

\section{Conclusion}

We argue that contrastive edits, which change the output of a \predictor to a given contrast prediction, are effective explanations of neural NLP models. We propose \textsc{Minimal Contrastive Editing} (\mice), a method for generating such edits. We introduce evaluation criteria for contrastive edits that are motivated by human contrastive explanations---{minimality} and {fluency}---and show that \mice edits for the \imdb, \newsgroups, and \race datasets are contrastive, fluent, and minimal. Through qualitative analysis of \mice edits, we show that they have utility for robust and reliable NLP system development.

\section{Broader Impact Statement}

\mice is intended to aid the interpretation of NLP models. As a model-agnostic explanation method, it has the potential to impact NLP system development across a wide range of models and tasks. In particular, \mice edits can benefit NLP model developers in facilitating debugging and exposing dataset artifacts, as discussed in \sect{sec:analysis}. As a consequence, they can also benefit downstream users of NLP models by facilitating access to less biased and more robust systems. 

While the focus of our work is on interpreting NLP models, there are potential misuses of \mice that involve other applications. Firstly, malicious actors might employ \mice to generate adversarial examples; for instance, they may aim to generate hate speech that is minimally edited such that it fools a toxic language classifier. Secondly, naively applying \mice for data augmentation could plausibly lead to less robust and more biased models: Because \mice edits are intended to expose issues in models, straightforwardly using them as additional training examples could reinforce existing artifacts and biases present in data. To mitigate this risk, we encourage researchers exploring data augmentation to carefully think about how to select and label edited instances.

We also encourage researchers to develop more efficient methods of generating minimal contrastive edits. As discussed in \sect{sec:discussion}, a limitation of \mice is its computational demand. Therefore, we recommend that future work focus on creating methods that require less compute.

\bibliography{acl2021}
\bibliographystyle{acl_natbib}

\clearpage

\appendix

\begin{table*}[t!]
\centering
\footnotesize
\begin{tabular}{M{0.05\linewidth}| m{0.44\linewidth} | m{0.41\linewidth}}\toprule
\normalsize\textbf{Task} & \multicolumn{1}{c|}{\normalsize\textbf{Original Input}} & \multicolumn{1}{c}{\normalsize\textbf{Input to \editor}}\\\midrule\midrule
\textsc{News} & Michael, you sent your inquiry to the bmw mailing list, but the sw replaces your return addr with the list addr so I can't reply or manually add you. please see my post re the list or contact me directly. & label: misc. input: <extra\_id\_0>, you sent your <extra\_id\_1> to the <extra\_id\_2>, but the <extra\_id\_3> your return <extra\_id\_4> with the list <extra\_id\_5> so I can't <extra\_id\_6> or <extra\_id\_7> add you. please see my post re the list or contact me directly.\\
\midrule
\race & \underline{article}: The best way of learning a language is by using it. The best way of learning English is using English as much as possible. Sometimes you will get your words mixed up and people won\'t understand. Sometimes people will say things too quickly and you can\'t understand them. But if you keep your sense of humor( ),you can always have a good laugh at the mistakes you make. Don\'t be unhappy if the people seem to  laugh at your mistakes. It\'s much better for people to laugh at your mistake than to be angry because they don\'t know what you are saying. The most important rule for learning English is "Don\'t be afraid of making mistakes. Everyone makes mistakes." \underline{question}: In learning English, you should \_. \underline{choices}: speak as quickly as possible., laugh as much as you can., use it as often as you can., write more than you read. & question: In learning English, you should \_. answer: choice1: laugh as much as you can. context: The <extra\_id\_0> <extra\_id\_1>. Sometimes you will get your words <extra\_id\_2> <extra\_id\_3> <extra\_id\_4> have a good laugh at the mistakes you make. Don't be unhappy if the people seem to  laugh at your mistakes. It's much better for people to laugh at your mistake than to be angry because they don't know what you are saying. The most important rule for learning English is "Don't be afraid of making mistakes. Everyone makes <extra\_id\_5>." choice0: speak as quickly as possible. choice1: laugh as much as you can. choice2: use it as often as you can. choice3: write more than you read.\\
\bottomrule
\end{tabular}
\caption{Examples of input formats to our \editors. 
The input to \newsgroups \editor has target label ``misc.''}
\label{tab:appendix_inputs}
\end{table*}
\begin{table*}[ht!]
\centering
\footnotesize
\begin{tabular}{m{0.95\linewidth}}\toprule
\multicolumn{1}{c}{\begin{tabular}{m{0.75\linewidth}} \vspace{-1mm}\begin{center}\hspace{1 mm}\textbf{Question:} \hspace{0.5 mm} \underline{\hspace{0.5cm}} of Xiao Maiyou's children went to Pecking University.\\
\hspace{0.55 mm} \hspace{0.5 mm} (a)\, {One} \, {(b) \,\, {Two}} \,(c) \,\, {Three} \, (d) \,\, \underline{All}\end{center}
\end{tabular}}\vspace{-3.5mm}\\
\begin{center}\textbf{Original pred} $y_p =$ (d)\, {\underline{All}} \hspace{11mm} \textbf{Contrast pred} $y_c =$ (b)\, {{Three}}\end{center} Just as "Tiger Mom" leaves, here comes the "Wolf Daddy" called Xiao Baiyou. He believes he's the best parent in the world. Some days ago, Xiao Baiyou's latest book about how to be a successful parent came out. He is pretty strict with his four children. Sometimes he even beat them. But the children don't hate their daddy at all. And all of them finally went to Pecking University\sout{,} \markedit{It is interesting to note that three of them got good marks at Pecking University. And} one of the \sout{top universities in China} \markedit{them even passed the exam without any problem}. So Xiao proudly tells others about his education idea that children need strict rules. In his microblog, he said, "Come on, want your children to enter Peking University without rules? You must be joking." And, "Leave your children more money, and strict rules at the same time."But the "Wolf Daddy" way was soon questioned by other parents. Some say that Xiao Baiyou just want to be famous by doing so. The "Wolf Daddy" Xiao Baiyou is a 47-year-old Guangdong businessman who deals in luxury goods in Hong Kong. Unlike many other parents who usually have one child, Xiao has four children. Two of them were born in Hong Kong and two in the US. Some people on the Internet think the reason why his children were able to enter Peking University is because the exam is much easier taken from Hong Kong.
\\

\bottomrule
\end{tabular}
\caption{A \mice edit for a prediction made by the ``buggy'' \race \predictor (described in \sect{sec:buggy_case_study}). Insertions are bolded in red. Deletions are struck through.
The true label for the original input is underlined.}
\label{tab:buggy_examples}
\end{table*}

\section{Training Details}

\subsection{\predictor Models}
\label{ssec:training-details}
For all datasets, $f$ is initialized as a \textsc{RoBERTa-large} model with a linear layer and maximum sequence length of 512 tokens. We train with \texttt{AllenNLP} \citep{Gardner2017AllenNLP}. For \imdb and \newsgroups, we fine-tune $f$ for 5 epochs with batch size 8 using Adam with initial learning rate of $2\mathrm{e}{-05}$, weight decay 0.1, and slanted triangular learning rate scheduler with cut frac 0.06. For \race, we fine-tune $f$ for 3 epochs with batch size 4 and 16 gradient accumulation steps using Adam with learning rate $1\mathrm{e}{-05}$, $\epsilon = 1\mathrm{e}{-08}$, and linear learning rate scheduler with 100 warm-up steps, and we fix $f$ after the epoch with the lowest validation loss.

\subsection{\editor Models}
\label{ssec:finetuning-details} We use the \texttt{transformers} implementation \citep{wolf-etal-2020-transformers} of the base \textsc{T5} for our \editors. We use Adam with a learning rate of $1\mathrm{e}{-4}$. For \imdb \editors, we use batch size 4 for all variants. For \newsgroups, we use batch size 4 for fine-tuning with predictor labels and batch size 8 for fine-tuning with gold labels. For \race, we use batch size 4 for fine-tuning with predictor labels and batch size 6 for fine-tuning with gold labels.

\section{Data Processing}
\label{sec:data-processing}

We remove newline and tab tokens (<br />, \textbackslash t, \textbackslash n) in all datasets, as these are tokenized differently by our \predictors (\textsc{RoBERTa-large}) and \editors (\textsc{T5}). For \newsgroups, we also remove headers, footers, and quotes.

\paragraph{Inputs to \editors} For \imdb and \newsgroups \editors, we simply prepend target labels to the masked original inputs. For \race, we give the question, context, all answer options, and the correct choice as input to the \race \editor. We only mask the context. See Table \ref{tab:appendix_inputs} for examples.

\section{T5 generation for large $n_2$}
\label{sec:appendix_t5_long}
We noticed that generations sometimes degenerate when we decode from T5 with a large masking percentage $n_2$. For example, sentinel tokens are sometimes generated out of consecutive order. We attribute this to the large difference between masking percentages we use (up to 55\%) and masking percentage used during T5 pretraining (15\%). Specifically, we observed that generations tend to degenerate after the the 28th sentinel token. Thus, we heuristically reduce the number of sentinel tokens by combining neighboring sentinel tokens that are separated by 1-2 tokens into one sentinel token.

When the output degenerates, we do the following: In-fill the mask tokens with the ``good'' parts of the generation (i.e.\ parts with correctly ordered sentinel tokens), and replace the remaining mask tokens with the original text; get the contrast label probabilities from $f$ for these intermediate in-filled candidates; of these, take the $m' = 3$ candidates with the highest probabilities and use as input to generate $m/m'$ new candidates.\footnote{ If one of the partially-infilled candidates results in the contrast label, we return this as the edited input.}

\section{Using \mice Edits to Debug a ``Buggy'' \predictor: A Case Study}
\label{sec:buggy_case_study}
In \sect{sec:analysis}, we illustrate how \mice edits can be used to debug both individual predictions and natural dataset artifacts learned by a model. Here, we further explore the utility of \mice edits in debugging through \textit{Data Staining} \cite{Sippy2020DataSA}: We design a ``buggy'' \predictor and evaluate whether \mice edits can recover the bug.

We create a buggy \race \predictor by introducing an artifact into the \race train set. This artifact is the presence of the phrase ``It is interesting to note that'' in front of the correct answer choice. We introduce this artifact as follows: We filter the \race train data to contain instances for which the correct answer choice is contained by some sentence\footnote{A sentence ``contains'' the correct answer choice if the answer has at least a 4-gram overlap with the sentence.} \textit{and} the overlapping sentence does not have a higher degree of n-gram overlap with some other (incorrect) choice. After filtering, 11,188 of 87,866 train instances remain. We then prepend ``It is interesting to note that'' to the overlapping sentence to design a correlation between the location of this phrase and the correct answer choice; our goal is to encourage a \predictor to learn to predict the multiple choice option closest to this buggy phrase as the correct answer. {If there are multiple overlapping sentences, we choose the one with the most overlap with the answer choice.} We randomly sample from this filtered subset such that 10\% of the train data contains this artifact. Our buggy \race \predictor is trained on this modified data using the same set-up from \sect{ssec:training-details}, except that we use a batch size of 2 and 32 gradient accumulation steps.

The test accuracies of our original and buggy \race \predictors are both 84\%, and so we cannot use this measure to select the better classifier. We ask whether \mice edits can be used for this purpose. One such edit is shown in Table \ref{tab:buggy_examples}. We observe that the signal from the edit, which contains both the manual artifact ``It is interesting to note that'' and the contrast prediction ``three,'' is enough to overpower the signal from the explicit assertion that ``All'' is the correct answer (``And all of them finally went to Pecking University'') such that the \predictor's prediction changes to ``Three.'' This edit thus provides evidence that some heuristic may have been learned by the predictor. Considering multiple \mice edits can validate such a hypothesis: We find that $17.2\%$ of the edits produced by \mice reflect this bug (i.e. contain the phrase ``interesting to note that''); in other words, they do uncover the manually inserted bug. 

Furthermore, \mice edits are able to uncover the artifact because they can \textit{insert} new text. For instance, in the edit in Table \ref{tab:appendix_inputs}, the buggy phrase ``It is interesting to note that'' is not part of the original input. Applying saliency-based explanation methods, such as gradient attribution, to the buggy \predictor's prediction would not reveal the \predictor's reliance on the manual artifact, as the buggy phrase is not already present in the text. This difference highlights a key advantage of \mice over existing instance-based explanation methods that attribute feature importance, which can only cite text already present in original inputs.


\begin{table*}[htbp]
\centering
\footnotesize
\begin{tabular}{m{0.95\linewidth}}\toprule
\multicolumn{1}{c}{\large\textbf{\imdb}}\\\midrule\midrule

\multicolumn{1}{c}{\begin{tabular}{m{0.75\linewidth}}\begin{center}\textbf{Original pred} $y_p=$ \,\underline{negative}\hspace{10 mm}\textbf{Contrast pred} $y_c=$ \,positive\end{center}\end{tabular}}\vspace{-2mm}\\
With a catchy title like the Butcher of Plainfield this Ed Gein variation and Kane Hodder playing him will no doubt fly off the shelves for a couple of weeks.Most viewers will be \sout{bored} \markedit{laughed} silly with this latest take on the life of Ed Gien. The movie focuses on Ed's rampage and gives us a(few)glimpses into his Psycosis and dwelling in Plainfeild.Its these scenes that give the movie a much needed jolt. \sout{What ruins this} \markedit{Another annoyance} is the constant focus on other characters lives and focuses less on Eds.Big mistake here. Kane Hodder is a strange choice to play Gein,but He does pull it off quite well,and deserves more acting credits than he gets these days.Prascilla Barnes and Micahel Barryman also show up. \sout{3/10} \markedit{9/10}\\

\midrule\multicolumn{1}{c}{\begin{tabular}{m{0.75\linewidth}} \begin{center}\textbf{Original pred} $y_p=$ \,\underline{positive} \hspace{10 mm}\textbf{Contrast pred} $y_c=$ \,negative\end{center}\end{tabular}}\vspace{-2mm}\\
I have just sat through this film again and can only wonder if we will see the \sout{likes} \markedit{kind} of films like this anymore? The \sout{timeless music} \markedit{sex}, the tender \sout{voices} \markedit{performances} of William Holden and Jennifer Jones leave this grown man \sout{weeping} \markedit{suffering} through \sout{joyous, romantic} \markedit{torturous, incoherent} scenes and I'm not one who cries very often in life. Where have our William Holden's gone and will they make these moving, \sout{wonderful} \markedit{cynical}, movies any more? It's sad to have to realize that they probably won't but don't think about it, just try to block that out of your mind. \sout{Even so} \markedit{Then again}, they won't have \sout{Holden} \markedit{Shakespeare} in it and he won't appear on that \sout{hill} \markedit{soap opera} just once more either. You can \sout{only enjoy} \markedit{safely skip} this film and watch it again.\\

\midrule
\multicolumn{1}{c}{\begin{tabular}{m{0.75\linewidth}} \begin{center}\textbf{Original pred} $y_p=$ \,\underline{positive} \hspace{10 mm}\textbf{Contrast pred} $y_c=$ \,negative\end{center}\end{tabular}}\vspace{-2mm}\\
This little flick is reminiscent of several other movies, but manages to keep its own style \& mood. "\sout{Troll} \markedit{Trusty}" \& "Don't Be Afraid of the Dark" come to mind. The \sout{suspense builders} \markedit{performances} were good, \& just cross the line from \sout{G} \markedit{silly} to \sout{PG} \markedit{uninteresting}. I especially liked the non-\sout{cliche} \markedit{cliched} choices with the parents; in other movies, I could predict the \sout{dialog} \markedit{ending} verbatim, but the writing in this movie made better selections. If you want a movie that's not \sout{gross} \markedit{terribly creepy} but gives you some chills, this is a great choice.\\

\bottomrule
\end{tabular}
\caption{Examples of edits produced by \mice for inputs from the \imdb dataset. Insertions are bolded in red. Deletions are struck through. $y_p$ is the \predictor's original prediction, and $y_c$ the contrast prediction. True labels for original inputs are underlined.}
\label{tab:imdb-examples-table}
\end{table*}

\begin{table*}[htbp]
\centering
\footnotesize
\begin{tabular}{m{0.95\linewidth}}\toprule
\multicolumn{1}{c|}{\large\textbf{\newsgroups}}\\\midrule\midrule

\multicolumn{1}{c}{\begin{tabular}{m{0.75\linewidth}} \begin{center}\textbf{Original pred} $y_p=$ \,\underline{talk}\hspace{10 mm}\textbf{Contrast pred} $y_c=$ \,sci\end{center}\end{tabular}}\vspace{-2mm}\\
Would someone be kind enought to document the exact nature of the evidence against the \sout{BD} \markedit{NRA}'s without reference to hearsay or newsreports. I would also like to know more about their past record etc. but again based on solid not media reports. My reason for asking for such evidence is that last night on Larry King Live a so-called "\sout{cult} \markedit{space}-expert" was interviewed from Australia who claimed that it was his evidence which led to the original \sout{raid} \markedit{discovery}. This admission, if true, raises the nasty possibility that the Government acted in good faith, which I believe they did, on faulty evidence. It also raises the possibility that other self proclaimed \sout{cult} \markedit{space} experts were advising them and giving ver poor advice.\\

\midrule

\multicolumn{1}{c}{\begin{tabular}{m{0.75\linewidth}} \begin{center}\textbf{Original pred} $y_p=$ \,\underline{rec}\hspace{10 mm}\textbf{Contrast pred} $y_c=$ \,soc\end{center}\end{tabular}}\vspace{-2mm}\\
I am planning a weekend in Chicago next month for my first live-and-in-person \sout{Cubs game} \markedit{Christian immersion} (!!!) I would appreciate any advice from locals or used-to-be locals on where to stay, what to see, where to dine, etc. E-mail replies are fine... Thanks in advance! Teresa\\

\midrule \multicolumn{1}{c}{\begin{tabular}{m{0.75\linewidth}} \begin{center}\textbf{Original pred} $y_p=$ \,\underline{rec}\hspace{10 mm}\textbf{Contrast pred} $y_c=$ \,alt\end{center}\end{tabular}}\vspace{-2mm}\\
\sout{Minor point: Shea Stadium} \markedit{(David: D.): This} was designed as a \sout{multi-purpose stadium} \markedit{symbiotic relationship between God- and-Christ} but not with the \sout{Jets in} \markedit{same} mind as the \sout{tennant} \markedit{Atheists}. The \sout{New York Football Giants} \markedit{Atheists} had moved to \sout{Yankee} \markedit{MetLife} Stadium (from the \sout{Polo Grounds} \markedit{Mets}) in \sout{1958} \markedit{1977} and was having problem with stadium management (the \sout{City} \markedit{Atheists} did not own \sout{Yankee} \markedit{MetLife} Stadium until \sout{1972} \markedit{1973}). The idea was to get the \sout{Giants} \markedit{Atheists} to move into \sout{Shea} \markedit{Metlife Stadium}. When a deal was worked out between the \sout{Giants} \markedit{Atheists} and the \sout{Yankees} \markedit{Mets,} the new \sout{AFL} \markedit{American} franchise, the \sout{New York Titans} \markedit{Atheists}, approached the \sout{City} \markedit{Mets} about using the new stadium. The \sout{Titans} \markedit{Mets} were playing in \sout{Downing} \markedit{Carling} Stadium (where the \sout{Cosmos} \markedit{Atheists} played \sout{soccer} \markedit{back} in the 70s). Because Shea Stadium was tied into the World's Fair anyway, the city thought it would be a novel idea to promote the new franchise and the World's Fair (like they were doing with the Mets). So the deal was worked out. I'm under the impression that when Murph says it, he means it! As a regular goer to Shea, it is not a bad place since they've cleaned and renovated the place. Remember, this is its 30th Year!\\

\bottomrule
\end{tabular}
\caption{Examples of edits produced by \mice for inputs from the \newsgroups dataset. Insertions are bolded in red. Deletions are struck through. $y_p$ is the \predictor's original prediction, and $y_c$ the contrast prediction. True labels for original inputs are underlined.}
\label{tab:news-examples-table}
\end{table*}
\begin{table*}[htbp]
\centering
\footnotesize
\begin{tabular}{m{0.95\linewidth}}\toprule
\multicolumn{1}{c|}{\large\textbf{\race}}\\\midrule\midrule

\multicolumn{1}{c}{\begin{tabular}{m{0.75\linewidth}} \hspace{15 mm}\textbf{Question:} \hspace{0.5 mm} How can the thieves get the information of the credit card?\\
\hspace{20 mm}(a)\, {The customers give them the information.}\\
\hspace{20 mm}(b)\, \underline{The thieves steal the information from Web sites.}\\
\hspace{20 mm}(c)\, {The customers sell the information to them.}\\
\hspace{20mm}(d)\, {The thieves buy the information from credit-card firms.}\\
\vspace{1mm}
\hspace{15 mm}\textbf{Original pred} $y_p=$ (a) \hspace{10 mm}\textbf{Contrast pred} $y_c=$ \underline{(b)}\end{tabular}}\vspace{-2mm}\\\\

The Internet has led to a huge increase in credit-card fraud. Your \sout{card} information could \sout{even} be for sale in an illegal web site. Web sites offering cheap goods and services should be regarded with care. On-line shoppers \sout{who enter} \markedit{can get credit-card information with stolen details through} their \sout{credit-card information may never receive the} \markedit{online shopping sites, including buying} goods they thought they bought. The thieves \sout{then go} \markedit{may use the information they have on your credit card to send} shopping \markedit{promotions, ads, or other Web sites. The thieves will not use} \sout{with} your card number -- or sell the information over the Internet. \sout{Computers} \markedit{Recent developments in internet} hackers have broken down security systems, raising questions about the safety of cardholder information. Several months ago, 25, 000 customers of CD Universe, an on-line music retailer, were not lucky. Their names, addresses and credit-card numbers were posted on a Web site after the retailer refused to pay US \$157, 828 to get back the information. Credit-card firms are now fighting against on-line fraud. Mastercard is working on plans for Web -- only credit card, with a lower credit limit. The card could be used only for shopping on-line \markedit{purchases}. \sout{However,} \markedit{But} there are a few simple steps you can take to keep from being cheated. Ask about your credit-card firm's on-line rules: Under British law, cardholders have to pay the first US \$ \sout{78}\markedit{20 penalty} of any fraudulent \sout{spending. And shop only} \markedit{activity} at secure sites; Send your credit-card information only if the Web site offers advanced secure system. If the security is in place, a letter will appear in the bottom right-hand corner of your screen. The Website address may also start https: //\sout{--} \markedit{// // // and}the extra "s" stands for secure. \sout{If in doubt,} \markedit{Never} give your credit-card information over the telephone. Keep your password safe: Most on-line sites require a user name and password \sout{before} \markedit{when} placing an order. Treat your passwords with care.\\

\midrule

\multicolumn{1}{c}{\begin{tabular}{m{0.75\linewidth}} \hspace{15 mm}\textbf{Question}: \hspace{0.5 mm} If you want to be a football player, you should \_\_.\\
\hspace{20 mm}(a)\, {buy a good football}\\
\hspace{20 mm}(b)\, \underline{play football}\\
\hspace{20 mm}(c)\, {watch others play football}\\
\hspace{20mm}(d)\, {put your football away}\\
\vspace{1mm}
\hspace{15 mm}\textbf{Original pred} $y_p =$ \underline{(b)} \hspace{10 mm}\textbf{Contrast pred} $y_c =$ (a)
\end{tabular}}\vspace{-2mm}\\\\
We are all learning English, but how can we learn English well? A student can know a lot about English, but maybe he can't speak English. If you want to \sout{know how to swim} \markedit{be a football player}, you must \sout{get into the river} \markedit{buy a good football. If} \sout{And if} you want to be \sout{a football} \markedit{an English} player, you must play football. So, you see. You can learn English only by using it. You must listen to your teacher in class. You must read your lessons every day. You must speak English to your classmates and also you must write something sometimes. Then one day, you may find your English very good.\\

\midrule

\multicolumn{1}{c}{\begin{tabular}{m{0.75\linewidth}} \hspace{15 mm}\textbf{Question}: \hspace{0.5 mm} This story most probably took place \_\_.\\
\hspace{20 mm}(a)\, {at the beginning of the term}\\
\hspace{20 mm}(b)\, {in the middle of the term}\\
\hspace{20 mm}(c)\, \underline{at the end of the term}\\
\hspace{20mm}(d)\, {at the beginning of the school year}\\
\vspace{1mm}
\hspace{15 mm}\textbf{Original pred} $y_p =$ \underline{(c)} \hspace{10 mm}\textbf{Contrast pred} $y_c=$ (b)
\end{tabular}}\vspace{-2mm}\\\\
A teacher \sout{stood} \markedit{was giving new classes to students} in \sout{front} \markedit{the middle} of \sout{his history} \markedit{this term. The students were in} class \sout{of twenty students just before handing out the final exam. His students} \markedit{by now. They} sat quietly and waited for him to speak. "It's been \sout{a pleasure teaching you this term} \markedit{my last chance}," he said \markedit{to them. The class started to cry. They cried for a long time. Finally, the teacher got up. He looked them in surprise. Then he asked them to leave. They} \sout{"You've} all \sout{worked very hard, so I have a pleasant surprise for you. Everyone who chooses not to take the final exam will get a 'B' for the course." Most of the students} jumped out of their seats. They thanked the teacher happily, and walked out of the classroom. Only a few students stayed. The teacher looked at them. "This is your last chance," he said. "Does anyone else want to leave?" All the students there stayed in their seats and took out their pencils. The teacher smiled. "Congratulations," he said. "I'm glad to see you believe in yourselves. You all get \sout{A} \markedit{on well}."\\

\bottomrule
\end{tabular}
\caption{Examples of edits produced by \mice for inputs from the \race dataset. Insertions are bolded in red. Deletions are struck through. $y_p$ is the \predictor's original prediction, and $y_c$ the contrast prediction. True labels for original inputs are underlined.}
\label{tab:race-examples-table}
\end{table*}

\end{document}